\documentclass[conference]{IEEEtran}
\IEEEoverridecommandlockouts
\usepackage{cite}
\usepackage{amsmath,amssymb,amsfonts}
\usepackage{algorithmic}
\usepackage{graphicx}
\usepackage{textcomp}
\usepackage{xcolor}
\usepackage{multicol}
\usepackage{graphicx} 
\usepackage{algorithm}
\usepackage{algorithmic}
\usepackage{float}

\usepackage{booktabs}
\def\BibTeX{{\rm B\kern-.05em{\sc i\kern-.025em b}\kern-.08em
    T\kern-.1667em\lower.7ex\hbox{E}\kern-.125emX}}

\title{Improving Drumming Robot Via Attention Transformer Network}

\makeatletter
\newcommand{\linebreakand}{%
  \end{@IEEEauthorhalign}
  \hfill\mbox{}\par
  \mbox{}\hfill\begin{@IEEEauthorhalign}
}
\makeatother

\author{
  \IEEEauthorblockN{Yang Yi}
  \IEEEauthorblockA{
    \textit{South China University of Technology}
    }
  \and
  
  \IEEEauthorblockN{Zonghan Li}
  \IEEEauthorblockA{
    \textit{South China University of Technology}
    }
   
}

\begin{document}
\maketitle

\begin{abstract}
Robotic technology has been widely used in nowadays society, which has made great progress in various fields such as agriculture, manufacturing and entertainment. In this paper, we focus on the topic of drumming robots in entertainment. To this end, we introduce an improving drumming robot that can automatically complete music transcription based on the popular vision transformer network based on the attention mechanism. Equipped with the attention transformer network, our method can efficiently handle the sequential audio embedding input and model their global long-range dependencies. Massive experimental results demonstrate that the improving algorithm can help the drumming robot promote drum classification performance, which can also help the robot to enjoy a variety of smart applications and services.

\end{abstract}


\section{Introduction}
Robotic technology, serves as an AIoT (Artificial Intelligence of Things)~\cite{ghosh2018artificial}, is widely used in recent scenes of our lives such as surveillance robot~\cite{su2020human,su2023improving}, navigation robot~\cite{desouza2002vision,gul2019comprehensive} and drumming robot~\cite{sui2020intelligent,yi2022aiot}. In this paper, we focus on the drumming robot for entertainment and try to solve the drum classification problem. To tackle such an issue, the key challenge is to transcribe the audio signals~\cite{wu2018review} to the regular sequential data structure for deep learning networks. And then design an efficient network for accurate classification. Based on the previous CNN-based drum transcription technique~\cite{yi2022aiot}, we introduce an improving drum transcription drum robot algorithm using the transformer network via the multi-head attention mechanism. Specifically, transformer network~\cite{dosovitskiy2020image} serves as the strong backbone to model the sequential embedding data and capture their global long-range semantic information, which can efficiently help the network to classify the music.

In general, our main contributions can be listed as follows:
\begin{itemize}
	\item [1)]
	We introduce a more powerful transformer network via an attention mechanism for drumming classification, which can significantly boost performance due to the ability of the network itself to handle sequential data. 
	\item [2)]
        We experimentally analyze the performance among different algorithms such as CNN, RNN and the transformer-based network. Through systemic evaluations, we reveal the advantage of the transformer network.
        \item [3)]
	Extensive experimental results demonstrate the effectiveness of the proposed drumming robot algorithm, which can achieve more competitive performance.
\end{itemize}

\section{Related Work}
\subsection{Robotic technology}
Previously, Sui et al.~\cite{sui2020intelligent} introduced an intelligent human interaction robot by using the traditional SVM classifier~\cite{joachims1998making} without any deep learning technique. Kotosaka et al.~\cite{kotosaka2001synchronized} proposed a framework of nonlinear oscillators for a robot system. Crick et al.~\cite{crick2006synchronization} also designed a multi-sensor data fusion method, including visual and auditory data, which enables a robot to drum in synchrony with human performers. In another study, Ince et al.~\cite{ince2015towards} presented a framework for drum stroke detection and recognition by using auditory cues. Based on turn-taking and imitation principles, they designed an interactive drumming game, in which the participants improved their ability to imitate by using the proposed framework. Li et al.~\cite{li2019light} later designed a light-weight convolutional network system for water meter reading in the smart city. Some recent robot systems~\cite{liu2019novel,zheng2018cognition} also try to combine edge-cloud~\cite{naveen2019key,yu2017survey} sides for data computing and data restoring. 

\begin{figure*}
	\begin{center}
		\centering
		\includegraphics[width=7.0in]{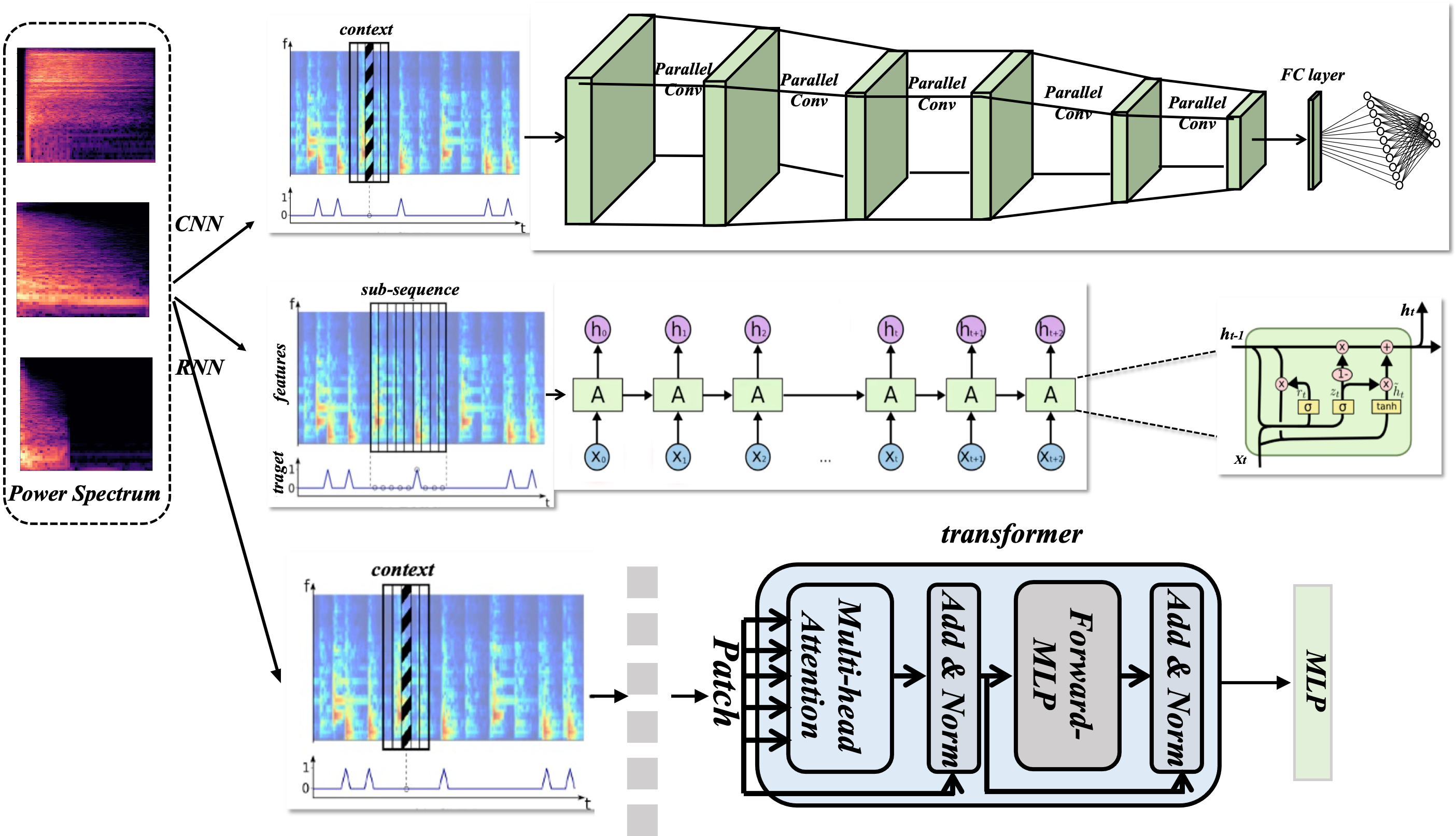}
	\end{center}
	\caption{An overview of drum transcription. We first extract the audio and convert it into the power-spectrogram, and then the network learns from the 2D image information. Specifically, we here compared three different deep neural networks: CNN, RNN and transformer.}
	\label{fig2}
\end{figure*}

\subsection{CNN-based Deep Learning Network}
CNN-based networks~\cite{chua1993cnn} are popular and have been applied for many tasks such as image classification~\cite{resnet1,vgg,imagenet}, image detection~\cite{su2022epnet,su2022self,tian2019fcos,SSD,yolo} and image segmentation~\cite{accuracy,dilatednet1,dilatednet2,su2021context}.
Due to the convenience and efficiency of convolutional operations, more and more works in other fields~\cite{yi2022aiot,sheykhivand2020recognizing,khamees2021classifying} utilize CNN to extract features from the input and then perform subsequent operations. 
However, CNN's local receptive field shortcomings limit its performance, and later some non-local~\cite{wang2018non,dai2017deformable} convolutional strategies are introduced.

\subsection{Transformer-based Deep Learning Network}
Transformer~\cite{vaswani2017attention} is first proposed in NLP~\cite{socher2012deep} area for processing sequential data. Later, vision transformer (ViT)~\cite{dosovitskiy2020image} is proposed for processing images or video. more recently, Vit has been applied for some other visual tasks such as editing~\cite{liu2021fuseformer}, low-level tasks~\cite{su2023unified,liang2021swinir} 3D reconstruction~\cite{wang2021multi} and so on. In general, the vision transformer explores the multi-head attention mechanism and reshapes the traditional grid input into token input, which can uniformly handle tasks in different fields and greatly accelerate the unified performance of deep learning.

\section{Methodology}

\subsection{Audio Extraction}
Since the raw input is the audio signal,l and thus, the input audio is firstly converted into a spectrogram using a short-time Fourier transform (STFT) with the help of librosa~\cite{mcfee2019librosa} using a Hanning window with a 2048 sample window size and 512 sample hop length. Then, the spectrogram is computed using a Mel-filter~\cite{farooq2001mel} in a frequency range of 20 to 20000 Hz with 128 Mel bands, resulting in a 128 $\times$ $n$ power-spectrogram. The whole pre-processing diagram strictly follows the previous work~\cite{yi2022aiot} for fair comparisons.

\subsection{Overall Architecture}
Fig~\ref{fig2} shows the overall architecture of our proposed method, which takes the converted 2D power-spectrogram audio data as input. Then, we encode the data by leveraging deep neural learning different networks such as CNN~\cite{chua1993cnn}, RNN~\cite{lipton2015critical} or the proposed transformer network.
Note that in our method, we use a transformer-tiny~\cite{dosovitskiy2020image} network as our backbone for efficient computing under limited resources.

\begin{table}[]
	\begin{center}\caption{\textbf{Table shows the results for different algorithms on drum transcription task.}}
	\label{table2}
		\begin{tabular}{cc}
			\hline
			Method & Accuracy(Top1\%)\\
			\hline\hline
			Sui $\emph{et~al.}$~\cite{sui2020intelligent} SVM &82.50 \\
			RNN &91.87 \\
			Yi $\emph{et~al.}$~\cite{yi2022aiot} & 92.18\\
                \textbf{Ours} &\textbf{93.68}\\
			\hline
		\end{tabular}
	\end{center}
\end{table}

\section{Experiments}
\textbf{Datasets:}  For the files of the drums audio, we collected various categories of sounds (Tom, Kick, Snare, Close-Hat, Ride, Crash and Open-hat) in the various open-source databases~\cite{dittmar2014real,gillet2006enst} on the Internet. Then, we divided them into the training sets and verification sets.

\textbf{Performance:} 
As shown in Table~\ref{table2}, we conduct systematic evaluations on the validation set for comparisons. We can observe that by adopting the transformer-based network, we can significantly boost the drumming classification performance and outperform the traditional SVM or RNN-based methods by a large margin. Besides, compared to the state-of-the-art CNN-based method~\cite{yi2022aiot}, we can also achieve competitive results, which reveals the superiority of the proposed network.

\section{Conclusion}
In this paper, we experimentally show the advantages of the proposed transformer-based drumming robot algorithm for music classification. By adopting the multi-head attention mechanism, our
network can achieve the new state-of-the-art performance. In future work, we will focus on designing more efficient algorithms such as some light-weight ViT~\cite{mehta2021mobilevit,li2022efficientformer} for network processing.

\bibliographystyle{IEEEtran}
\bibliography{ref}

\begin{thebibliography}{10}
\providecommand{\url}[1]{#1}
\csname url@samestyle\endcsname
\providecommand{\newblock}{\relax}
\providecommand{\bibinfo}[2]{#2}
\providecommand{\BIBentrySTDinterwordspacing}{\spaceskip=0pt\relax}
\providecommand{\BIBentryALTinterwordstretchfactor}{4}
\providecommand{\BIBentryALTinterwordspacing}{\spaceskip=\fontdimen2\font plus
\BIBentryALTinterwordstretchfactor\fontdimen3\font minus
  \fontdimen4\font\relax}
\providecommand{\BIBforeignlanguage}[2]{{%
\expandafter\ifx\csname l@#1\endcsname\relax
\typeout{** WARNING: IEEEtran.bst: No hyphenation pattern has been}%
\typeout{** loaded for the language `#1'. Using the pattern for}%
\typeout{** the default language instead.}%
\else
\language=\csname l@#1\endcsname
\fi
#2}}
\providecommand{\BIBdecl}{\relax}
\BIBdecl

\bibitem{ghosh2018artificial}
A.~Ghosh, D.~Chakraborty, and A.~Law, ``Artificial intelligence in internet of
  things,'' \emph{CAAI Transactions on Intelligence Technology}, vol.~3, no.~4,
  pp. 208--218, 2018.

\bibitem{su2020human}
Y.~Su, G.~Lin, J.~Zhu, and Q.~Wu, ``Human interaction learning on 3d skeleton
  point clouds for video violence recognition,'' in \emph{Computer Vision--ECCV
  2020: 16th European Conference, Glasgow, UK, August 23--28, 2020,
  Proceedings, Part IV 16}.\hskip 1em plus 0.5em minus 0.4em\relax Springer,
  2020, pp. 74--90.

\bibitem{su2023improving}
Y.~Su, G.~Lin, and Q.~Wu, ``Improving video violence recognition with human
  interaction learning on 3d skeleton point clouds,'' \emph{arXiv preprint
  arXiv:2308.13866}, 2023.

\bibitem{desouza2002vision}
G.~N. DeSouza and A.~C. Kak, ``Vision for mobile robot navigation: A survey,''
  \emph{IEEE transactions on pattern analysis and machine intelligence},
  vol.~24, no.~2, pp. 237--267, 2002.

\bibitem{gul2019comprehensive}
F.~Gul, W.~Rahiman, and S.~S. Nazli~Alhady, ``A comprehensive study for robot
  navigation techniques,'' \emph{Cogent Engineering}, vol.~6, no.~1, p.
  1632046, 2019.

\bibitem{sui2020intelligent}
L.~Sui, Y.~Su, Y.~Yi, Z.~Li, and J.~Zhu, ``Intelligent drumming robot for human
  interaction,'' in \emph{2020 International Symposium on Autonomous Systems
  (ISAS)}.\hskip 1em plus 0.5em minus 0.4em\relax IEEE, 2020, pp. 168--173.

\bibitem{yi2022aiot}
Y.~Yi, M.~Lu, L.~Wu, and Z.~Chen, ``Aiot-based drum transcription robot using
  convolutional neural networks,'' in \emph{4th International Conference on
  Informatics Engineering \& Information Science (ICIEIS2021)}, vol.
  12161.\hskip 1em plus 0.5em minus 0.4em\relax SPIE, 2022, pp. 103--108.

\bibitem{wu2018review}
C.-W. Wu, C.~Dittmar, C.~Southall, R.~Vogl, G.~Widmer, J.~Hockman,
  M.~M{\"u}ller, and A.~Lerch, ``A review of automatic drum transcription,''
  \emph{IEEE/ACM Transactions on Audio, Speech, and Language Processing},
  vol.~26, no.~9, pp. 1457--1483, 2018.

\bibitem{dosovitskiy2020image}
A.~Dosovitskiy, L.~Beyer, A.~Kolesnikov, D.~Weissenborn, X.~Zhai,
  T.~Unterthiner, M.~Dehghani, M.~Minderer, G.~Heigold, S.~Gelly \emph{et~al.},
  ``An image is worth 16x16 words: Transformers for image recognition at
  scale,'' \emph{arXiv preprint arXiv:2010.11929}, 2020.

\bibitem{joachims1998making}
T.~Joachims, ``Making large-scale svm learning practical,'' Technical report,
  Tech. Rep., 1998.

\bibitem{kotosaka2001synchronized}
S.~Kotosaka and S.~Schaal, ``Synchronized robot drumming by neural
  oscillator,'' \emph{Journal of the Robotics Society of Japan}, vol.~19,
  no.~1, pp. 116--123, 2001.

\bibitem{crick2006synchronization}
C.~Crick, M.~Munz, and B.~Scassellati, ``Synchronization in social tasks:
  Robotic drumming,'' in \emph{ROMAN 2006-The 15th IEEE International Symposium
  on Robot and Human Interactive Communication}.\hskip 1em plus 0.5em minus
  0.4em\relax IEEE, 2006, pp. 97--102.

\bibitem{ince2015towards}
G.~Ince, T.~B. Duman, R.~Yorganci, and H.~Kose, ``Towards a robust drum stroke
  recognition system for human robot interaction,'' in \emph{2015 IEEE/SICE
  International Symposium on System Integration (SII)}.\hskip 1em plus 0.5em
  minus 0.4em\relax IEEE, 2015, pp. 744--749.

\bibitem{li2019light}
C.~Li, Y.~Su, R.~Yuan, D.~Chu, and J.~Zhu, ``Light-weight spliced convolution
  network-based automatic water meter reading in smart city,'' \emph{IEEE
  Access}, vol.~7, pp. 174\,359--174\,367, 2019.

\bibitem{liu2019novel}
J.~Liu, F.~Zhou, L.~Yin, and Y.~Wang, ``A novel cloud platform for service
  robots,'' \emph{IEEE Access}, 2019.

\bibitem{zheng2018cognition}
J.~Zheng, Q.~Zhang, S.~Xu, H.~Peng, and Q.~Wu, ``Cognition-based context-aware
  cloud computing for intelligent robotic systems in mobile education,''
  \emph{IEEE Access}, vol.~6, pp. 49\,103--49\,111, 2018.

\bibitem{naveen2019key}
S.~Naveen and M.~R. Kounte, ``Key technologies and challenges in iot edge
  computing,'' in \emph{2019 Third international conference on I-SMAC (IoT in
  social, mobile, analytics and cloud)(I-SMAC)}.\hskip 1em plus 0.5em minus
  0.4em\relax IEEE, 2019, pp. 61--65.

\bibitem{yu2017survey}
W.~Yu, F.~Liang, X.~He, W.~G. Hatcher, C.~Lu, J.~Lin, and X.~Yang, ``A survey
  on the edge computing for the internet of things,'' \emph{IEEE access},
  vol.~6, pp. 6900--6919, 2017.

\bibitem{chua1993cnn}
L.~O. Chua and T.~Roska, ``The cnn paradigm,'' \emph{IEEE Transactions on
  Circuits and Systems I: Fundamental Theory and Applications}, vol.~40, no.~3,
  pp. 147--156, 1993.

\bibitem{resnet1}
K.~He, X.~Zhang, S.~Ren, and J.~Sun, ``Deep residual learning for image
  recognition,'' in \emph{Proceedings of the IEEE conference on computer vision
  and pattern recognition}, 2016, pp. 770--778.

\bibitem{vgg}
K.~Simonyan and A.~Zisserman, ``Very deep convolutional networks for
  large-scale image recognition,'' \emph{arXiv preprint arXiv:1409.1556}, 2014.

\bibitem{imagenet}
O.~Russakovsky, J.~Deng, H.~Su, J.~Krause, S.~Satheesh, S.~Ma, Z.~Huang,
  A.~Karpathy, A.~Khosla, M.~Bernstein \emph{et~al.}, ``Imagenet large scale
  visual recognition challenge,'' \emph{International journal of computer
  vision}, vol. 115, no.~3, pp. 211--252, 2015.

\bibitem{su2022epnet}
J.~Su, Y.~Su, Y.~Zhang, W.~Yang, H.~Huang, and Q.~Wu, ``Epnet: Power lines
  foreign object detection with edge proposal network and data composition,''
  \emph{Knowledge-Based Systems}, vol. 249, p. 108857, 2022.

\bibitem{su2022self}
Y.~Su, G.~Lin, Y.~Hao, Y.~Cao, W.~Wang, and Q.~Wu, ``Self-supervised object
  localization with joint graph partition,'' in \emph{Proceedings of the AAAI
  Conference on Artificial Intelligence}, vol.~36, no.~2, 2022, pp. 2289--2297.

\bibitem{tian2019fcos}
Z.~Tian, C.~Shen, H.~Chen, and T.~He, ``Fcos: Fully convolutional one-stage
  object detection,'' in \emph{Proceedings of the IEEE/CVF international
  conference on computer vision}, 2019, pp. 9627--9636.

\bibitem{SSD}
W.~Liu, D.~Anguelov, D.~Erhan, C.~Szegedy, S.~Reed, C.-Y. Fu, and A.~C. Berg,
  ``Ssd: Single shot multibox detector,'' in \emph{European conference on
  computer vision}.\hskip 1em plus 0.5em minus 0.4em\relax Springer, 2016, pp.
  21--37.

\bibitem{yolo}
J.~Redmon, S.~Divvala, R.~Girshick, and A.~Farhadi, ``You only look once:
  Unified, real-time object detection,'' in \emph{Proceedings of the IEEE
  conference on computer vision and pattern recognition}, 2016, pp. 779--788.

\bibitem{accuracy}
J.~Long, E.~Shelhamer, and T.~Darrell, ``Fully convolutional networks for
  semantic segmentation,'' in \emph{Proceedings of the IEEE conference on
  computer vision and pattern recognition}, 2015, pp. 3431--3440.

\bibitem{dilatednet1}
L.-C. Chen, G.~Papandreou, I.~Kokkinos, K.~Murphy, and A.~L. Yuille, ``Semantic
  image segmentation with deep convolutional nets and fully connected crfs,''
  \emph{arXiv preprint arXiv:1412.7062}, 2014.

\bibitem{dilatednet2}
F.~Yu and V.~Koltun, ``Multi-scale context aggregation by dilated
  convolutions,'' \emph{arXiv preprint arXiv:1511.07122}, 2015.

\bibitem{su2021context}
Y.~Su, R.~Sun, G.~Lin, and Q.~Wu, ``Context decoupling augmentation for weakly
  supervised semantic segmentation,'' in \emph{Proceedings of the IEEE/CVF
  international conference on computer vision}, 2021, pp. 7004--7014.

\bibitem{sheykhivand2020recognizing}
S.~Sheykhivand, Z.~Mousavi, T.~Y. Rezaii, and A.~Farzamnia, ``Recognizing
  emotions evoked by music using cnn-lstm networks on eeg signals,'' \emph{IEEE
  access}, vol.~8, pp. 139\,332--139\,345, 2020.

\bibitem{khamees2021classifying}
A.~A. Khamees, H.~D. Hejazi, M.~Alshurideh, and S.~A. Salloum, ``Classifying
  audio music genres using cnn and rnn,'' in \emph{International Conference on
  Advanced Machine Learning Technologies and Applications}.\hskip 1em plus
  0.5em minus 0.4em\relax Springer, 2021, pp. 315--323.

\bibitem{wang2018non}
X.~Wang, R.~Girshick, A.~Gupta, and K.~He, ``Non-local neural networks,'' in
  \emph{Proceedings of the IEEE conference on computer vision and pattern
  recognition}, 2018, pp. 7794--7803.

\bibitem{dai2017deformable}
J.~Dai, H.~Qi, Y.~Xiong, Y.~Li, G.~Zhang, H.~Hu, and Y.~Wei, ``Deformable
  convolutional networks,'' in \emph{Proceedings of the IEEE international
  conference on computer vision}, 2017, pp. 764--773.

\bibitem{vaswani2017attention}
A.~Vaswani, N.~Shazeer, N.~Parmar, J.~Uszkoreit, L.~Jones, A.~N. Gomez,
  {\L}.~Kaiser, and I.~Polosukhin, ``Attention is all you need,''
  \emph{Advances in neural information processing systems}, vol.~30, 2017.

\bibitem{socher2012deep}
R.~Socher, Y.~Bengio, and C.~D. Manning, ``Deep learning for nlp (without
  magic),'' in \emph{Tutorial Abstracts of ACL 2012}, 2012, pp. 5--5.

\bibitem{liu2021fuseformer}
R.~Liu, H.~Deng, Y.~Huang, X.~Shi, L.~Lu, W.~Sun, X.~Wang, J.~Dai, and H.~Li,
  ``Fuseformer: Fusing fine-grained information in transformers for video
  inpainting,'' in \emph{Proceedings of the IEEE/CVF international conference
  on computer vision}, 2021, pp. 14\,040--14\,049.

\bibitem{su2023unified}
Y.~Su, J.~Deng, R.~Sun, G.~Lin, H.~Su, and Q.~Wu, ``A unified transformer
  framework for group-based segmentation: Co-segmentation, co-saliency
  detection and video salient object detection,'' \emph{IEEE Transactions on
  Multimedia}, 2023.

\bibitem{liang2021swinir}
J.~Liang, J.~Cao, G.~Sun, K.~Zhang, L.~Van~Gool, and R.~Timofte, ``Swinir:
  Image restoration using swin transformer,'' in \emph{Proceedings of the
  IEEE/CVF international conference on computer vision}, 2021, pp. 1833--1844.

\bibitem{wang2021multi}
D.~Wang, X.~Cui, X.~Chen, Z.~Zou, T.~Shi, S.~Salcudean, Z.~J. Wang, and
  R.~Ward, ``Multi-view 3d reconstruction with transformers,'' in
  \emph{Proceedings of the IEEE/CVF International Conference on Computer
  Vision}, 2021, pp. 5722--5731.

\bibitem{mcfee2019librosa}
B.~McFee, V.~Lostanlen, M.~McVicar, A.~Metsai, S.~Balke, C.~Thom{\'e},
  C.~Raffel, A.~Malek, D.~Lee, F.~Zalkow \emph{et~al.}, ``librosa/librosa: 0.7.
  2,'' \emph{Version 0.7}, vol.~1, 2019.

\bibitem{farooq2001mel}
O.~Farooq and S.~Datta, ``Mel filter-like admissible wavelet packet structure
  for speech recognition,'' \emph{IEEE signal processing letters}, vol.~8,
  no.~7, pp. 196--198, 2001.

\bibitem{lipton2015critical}
Z.~C. Lipton, J.~Berkowitz, and C.~Elkan, ``A critical review of recurrent
  neural networks for sequence learning,'' \emph{arXiv preprint
  arXiv:1506.00019}, 2015.

\bibitem{dittmar2014real}
C.~Dittmar and D.~G{\"a}rtner, ``Real-time transcription and separation of drum
  recordings based on nmf decomposition.'' in \emph{DAFx}, 2014, pp. 187--194.

\bibitem{gillet2006enst}
O.~Gillet and G.~Richard, ``Enst-drums: an extensive audio-visual database for
  drum signals processing.'' in \emph{ISMIR}, 2006, pp. 156--159.

\bibitem{mehta2021mobilevit}
S.~Mehta and M.~Rastegari, ``Mobilevit: light-weight, general-purpose, and
  mobile-friendly vision transformer,'' \emph{arXiv preprint arXiv:2110.02178},
  2021.

\bibitem{li2022efficientformer}
Y.~Li, G.~Yuan, Y.~Wen, J.~Hu, G.~Evangelidis, S.~Tulyakov, Y.~Wang, and
  J.~Ren, ``Efficientformer: Vision transformers at mobilenet speed,''
  \emph{Advances in Neural Information Processing Systems}, vol.~35, pp.
  12\,934--12\,949, 2022.

\end{thebibliography}

\end{document}